# Enhancing Omics Cohort Discovery for Research on Neurodegeneration through Ontology-Augmented Embedding Models

José A. Pardo[1], Alicia Gómez-Pascual[1], José T. Palma[1], Juan A. Botía[1,2*]

[1]Department of Information and Communications Engineering, University of Murcia.
[2]Department of Neurodegenerative diseases, University College London.

*Corresponding author: juanbot@um.es

## Abstract
The growing volume of omics and clinical data generated for neurodegenerative diseases (NDs) requires new approaches for their curation so they can be ready-to-use in bioinformatics. NeuroEmbed is an approach for the engineering of semantically accurate embedding spaces to represent cohorts and samples. The NeuroEmbed method comprises four stages: (1) extraction of ND cohorts from public repositories; (2) semi-automated normalization and augmentation of metadata of cohorts and samples using biomedical ontologies and clustering on the embedding space; (3) automated generation of a natural language question-answering (QA) dataset for cohorts and samples based on randomized combinations of standardized metadata dimensions and (4) fine-tuning of a domain-specific embedder to optimize queries. We illustrate the approach using the GEO repository and the PubMedBERT pretrained embedder. Applying NeuroEmbed, we semantically indexed 2,801 repositories and 150,924 samples. Amongst many biology-relevant categories, we normalized more than 1,700 heterogeneous tissue labels from GEO into 326 unique ontology-aligned concepts and enriched annotations with new ontology-aligned terms, leading to a fold increase in size for the metadata terms between 2.7 and 20 fold. After fine-tuning PubMedBERT with the QA training data augmented with the enlarged metadata, the model increased its mean Retrieval Precision from 0.277 to 0.866 and its mean Percentile Rank from 0.355 to 0.896. The NeuroEmbed methodology for the creation of electronic catalogues of omics cohorts and samples will foster automated bioinformatic pipelines construction. The NeuroEmbed catalogue of cohorts and samples is available at https://github.com/JoseAdrian3/NeuroEmbed.

## Introduction
In the past decade, public repositories (e.g., Gene Expression Omnibus (GEO) [1], ArrayExpress [2], European Genome-phenome Archive (EGA) [3], Accelerating

Medicins Partnership Parkinson's Disease (AMP-PD) [4], Synapse [5]) have facilitated access to thousands of omics datasets accompanied by clinical metadata. In this context, metadata refers to structured or unstructured information describing or providing context for biological datasets (e.g., clinical annotation, study protocol, patient demographics…). Over 80% of this companion clinical metadata is still stored as free text or within heterogeneous containers [6]. Furthermore, the absence of standardized terminologies fosters redundancy (e.g., "UPDRS-III" and "motor score"), which hinders the ability to identify specific concepts (semantic precision), as well as the ability to detect the same concepts across different cohorts (semantic coherence). This largely heterogeneous and weakly structured metadata limits the retrieval and comparison of relevant cohorts across studies [1].

Multiple reviews propose harmonization of formats (e.g, ontologies) as a partial solution to these limitations. However, current methodological guidance remains insufficient to support automated integration of unstructured or semi-structured biological data into structured repositories. [7,8].

Knowledge modeling consists of turning free-text, highly heterogeneous descriptions of relevant clinical information into structured knowledge. It addresses simultaneously the task of acquiring information and organising it in a more or less formal structure, e.g., using an ontology as a reference [4]. Early approaches relied on rule-based or shallow NLP, achieving only incremental fact extraction [9,10]; the advent of Large Language Models (LLM) is significantly accelerating the conversion of unstructured information metadata into much more useful and structured resources. Recent work has explored the use of LLMs to automate the formal arrangement of clinical metadata. LLM-based approaches for knowledge extraction are able to recover high-accuracy entities from unstructured health records, achieving remarkable high performance [11]. This in combination with ontologies as the reference framework to be used as knowledge structure is the most successful approach to date. Besides, the combination of different ontologies enhances the prediction of biomedical associations and improves cohort search capabilities by providing cross-domain semantic terms [11,12]. Hybrid strategies that combine knowledge-graph structures with contrastive learning mechanisms have recently shown superior performance in retrieving biomedical cohorts compared to single-source or non-semantic baselines [13]. These methods construct dense representations informed by ontological hierarchies, thereby increasing both retrieval precision and robustness across domains [13].

Unlike prior hybrid methods, our approach targets cohort retrieval in the domain of ND research, including Alzheimer's disease (AD), Parkinson's disease (PD), Huntington's disease (HD), Lewy Body Disease (LWB), Frontotemporal Dementia (FT) and Multiple

System Atrophy (MSA). It can also be adapted to any domain, as long as there is a structured body of knowledge that can be used as reference. Additionally, our method integrates external domain knowledge using a two steps approach. First, we expanded clinical metadata using biomedical ontologies knowledge, allowing the model to disambiguate terms and align related concepts. Then, we train the model to recognize semantic similarities and differences in the data. This is done through a learning technique (contrastive learning) that pulls together the representations of conceptually related inputs and pushes apart unrelated ones in the embedding space, ensuring that similar clinical terms are embedded close to each other. By integrating these two techniques, we introduce a model that produces embeddings for clinical data that are augmented with the biomedical-ontology-based information. These ontology-augmented embeddings are designed to enhance the model's ability to retrieve meaningful cohorts from complex biomedical repositories [14]. We define four key metadata dimensions to describe each ND cohort: Population (Po), Assay (As), Phenotype (Ph), and Tissue (Ti). Applied to the GEO repository and grounded on PubMedBERT embedder, our framework generates embeddings that reconcile heterogeneous clinical descriptors within a unified semantic space, enabling precise and scalable retrieval of ND cohorts and paving the way for more comprehensive multi-omics analyses.

## Methodology

We implemented a six-step pipeline that systematically transforms the heterogeneous, unstructured metadata of the ND studies into a semantically enriched, query-ready repository, publicly accessible to any user (see Figure 1): (0) acquisition of ND cohorts using Medical Subject Headings (MeSH) queries (e.g., "Parkinson") [15]; (1) generation of synonyms for the key metadata dimensions (Po, As, Ph, Ti) using biomedical ontologies; (2) generation of a Natural Language Queries Question Answering (NLQ-QA) dataset by combining original and synonym metadata and linking them to specific cohorts; (4) fine-tuning of a PubMedBERT-based embedding model on the QA dataset; (5) evaluation of the embedding's performance using standard retrieval metrics; and (6) deployment of the final embeddings to enable semantic search on the enriched metadata.

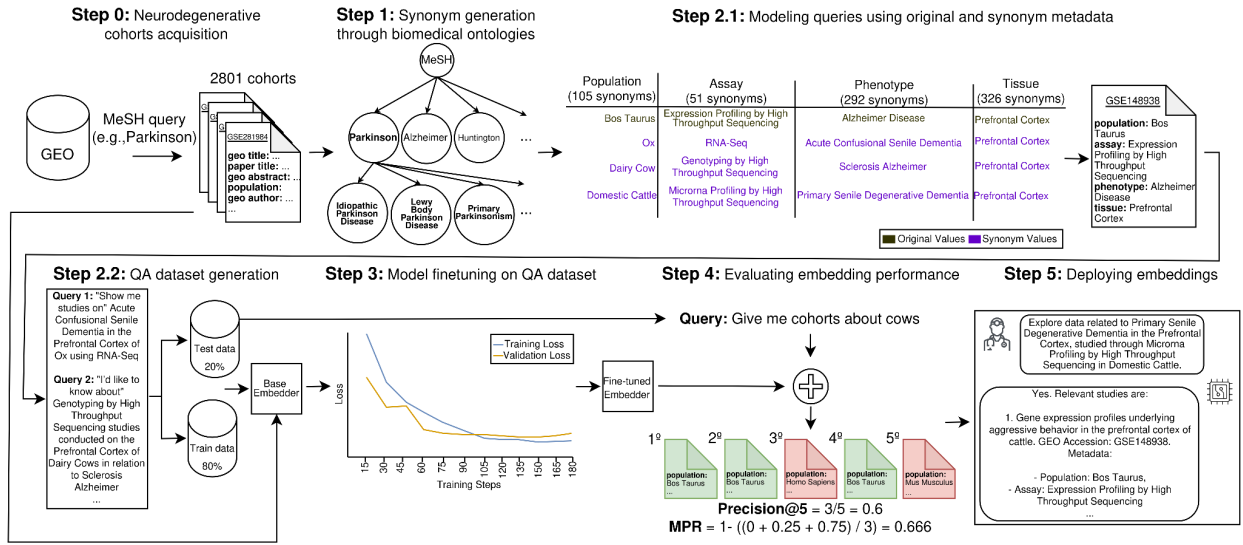

**Figure 1. Workflow for the discovery of neurodegenerative cohorts through ontology-augmented embedding. Step 0:** Retrieval of GEO neurodegenerative cohorts using MeSH queries (e.g., Parkinson). **Step 1:** Synonym augmentation via biomedical ontologies (e.g., MeSH) across all four dimensions. **Step 2:** Query modeling through substitution of original metadata with ontology-augmented synonyms and linking queries to cohort entries. **Step 3:** Generation of QA dataset with NLQ and split into training and test sets. **Step 4:** Fine-tuning of a base biomedical embedder and monitoring training and evaluation loss. **Step 5:** Evaluation using Precision and MPR metrics. **Step 6:** Deployment of the fine-tuned model through an interactive interface.

### Step 0: data acquisition

To address disease-specific heterogeneity across NDs, we systematically collected all studies of the most prevalent NDs that included any type of omics data and basic clinical and demographic data of the participants from the GEO repository (Figure 1, step 0). To identify disease-specific cohorts with a high confidence, we constructed disease-specific queries using MeSH medical thesaurus terms and their synonyms , (e.g., "Idiopathic Parkinson's Disease" for "Parkinson's disease"), since GEO does not impose rigorous metadata specifications on the studies.. To controlling the false positives cohorts, we applied a filtering protocol that requires at least one disease-specific MeSH term or synonym to appear in any of the following data fields: (1) GEO study title, (2) publication title (if available), (3) GEO abstract, excluding the first and last sentences to avoid boilerplate or non-informative content. Additionally, we used the *pymed* Python library [16] to retrieve publication-level metadata (e.g., article's title), since GEO only provides the study metadata and PubMed ID (PMID)

### Step 1: synonym generation

After identifying relevant cohorts, we applied a synonym search protocol to normalize and expand the range of related terms and capture more semantic and enriched representations of the metadata dimensions (see Figure 1, Step 1). These ontology-based searches allow us to: (1) normalized the data, i.e. consolidating those variants into a single canonical identifier so that semantically equivalent concepts are represented consistently across cohorts; (2) augmentate the data; i.e. systematic expansion of each original metadata term with all ontology-derived synonym variants.

The augmentation process followed a two-stage matching strategy. First, we searched for exact matches of values from the metadata at the target ontologies. Those found at any of the ontologies were used to expand with synonyms. Ontologies included the Experimental Factor Ontology (EFO) [17] for assay related terms, UBERON [18] for tissue terms and NCBI Taxonomy [19,20] for population values. If no match was found in these primary ontologies, the search extended to broader ontologies, i.e., MeSH [15], followed by UMLS [21,22]. OWL ontologies (EFO, NCBI Taxonomy, UBERON) were queried via the "hasExactSynonym" field using the "rdflib" [23] and "xml" [24] libraries. MeSH synonyms were extracted from the "ConceptList" function. UMLS terms were mapped using a precompiled Concept Unique Identifier dictionary. If no exact match was found for a value, either a population, assay, phenotype or tissue, then a fuzzy string matching was performed using the "thefuzz" Python library [23,25] where only candidates with a similarity score >= 80% based on Levenshtein distance [14].

**Step 2: QA dataset generation**

After metadata augmentation, we leveraged all these base metadata to generate NLQs related to our cohorts (see Figure 1, Step 2). Each query was constructed by randomly combining one to four original or augmented values from the four metadata dimensions. Only NLQs for which at least one matching cohort was available in our cohort dataset were retained. For example, an NLQ could be: *Show me cohorts within ox population from prefrontal cortex from transcription profiling by high throughput sequencing with senile dementia alzheimer type observations*, where the four metadata dimensions (Po, Ti, As and Ph) are queried. Another example could be: *I'm searching for cohorts with ataxia limb from rna profiling by array assay*, where just two dimensions (Phenotype and Assay) are combined.

To generate the Question Answering Dataset (QAD), we performed a stratified split of the synonym-expanded vocabulary of the four cohort-metadata dimensions to get exclusive training and test synonym sets in a 80/20 ratio (Algorithm 1 Line 2). For every query (i.e. the combination of one to four training values), we retrieved the cohorts whose metadata satisfied the entire combination and randomly selected one of them (Algorithm 1 Lines 5–11). Each pair of query-cohort was then converted into a NLQ by using one of six predefined templates (one of them is exclusive for test) and adding prepositions to make the queries more real. The predefined templates used were: (1) "Give me papers about…"; (2) "Can you show findings about…"; (3) "Explore data related to…"; (4) "Show me studies on…"; (5) "What research exists on…", and (6) "I'd like to know about…". Finally, we formed the training set with NLQ-cohort pairs that use only training synonyms and five of the six templates, and the test set with only test synonyms (Algorithm 1 Lines 20–22). This dual partitioning allowed us to evaluate the model's capacity to generalize across both (1) novel synonym instances and (2) unseen query formulations.

---

**Require:** $vocab\_augmented$, $cohorts$, $templates$, $prefix\_suffix\_rules$

1: **procedure** GENERATEQAD
2:    $(train\_vals, test\_vals) \leftarrow$ STRATIFIEDSPLIT($vocab\_augmented$, 0.8)
3:    $pair\_list \leftarrow \{\}$
4:    $final\_QA \leftarrow \{\}$

5:    **for** $k \leftarrow 1$ **to** 4 **do**
6:       **for all** $combo \in$ RANDOMCOMBINATIONS($train\_vals$, $k$) **do**
7:          $compatible \leftarrow$ FILTER($cohorts$, $combo$)
8:          **if** $compatible \neq \varnothing$ **then**
9:             $cohort \leftarrow$ RANDOMCHOICE($compatible$)
10:            $pair\_list \leftarrow pair\_list \cup \{(q, cohort)\}$
11:          **end if**
12:       **end for**
13:    **end for**

14:    **for all** $(q, cohort) \in pair\_list$ **do**
15:       $variants \leftarrow$ CREATENLQ($q$, $templates$, $prefix\_suffix\_rules$)
16:       **for all** $nlq \in variants$ **do**
17:          $final\_QA \leftarrow final\_QA \cup \{(nlq, cohort)\}$
18:       **end for**
19:    **end for**

20:    $train\_set \leftarrow \{p \in final\_QA \mid$ USESONLY($train\_vals$, $p.nlq$) $\land$ $\neg$TEMPLATEISTESTONLY($p.nlq$)$\}$
21:    $test\_set \leftarrow \{p \in final\_QA \mid$ USESONLY($test\_vals$, $p.nlq$)$\}$
22:    **return** ($train\_set$, $test\_set$)
23: **end procedure**

---

**Algorithm 1. Generation of QA training dataset from ontology-augmented metadata dimensions.** The procedure takes as input the augmented vocabulary across four metadata dimensions (Po, As, Ph, Ti) and follows these steps: (1) Vocabulary is split into training and testing subsets using a stratified strategy (line 2); (2) For each combination of 1 to 4 training terms, cohorts satisfying the corresponding constraints are filtered, and if at least one valid cohort exists, a query–cohort pair is stored (lines 5–13); (3) For each query–cohort pair, NLQs are generated using predefined templates and prefix/suffix rules (lines 14–19); (4) The resulting NLQs are filtered into a training set and a test set depending on which subset their terms belong to and whether the template is marked as test-only (lines 20–22). This process yields a balanced and diverse QA dataset suitable for fine-tuning natural language embedders.

**Step 3: model fine-tuning**

To adapt a biomedical language embedder to the constructed QAD, we fine-tuned the NeuML/pubmedbert-base-embeddings model [25] (see Figure 1, Step 5), pretrained exclusively on 14 million PubMed abstracts and full-text articles, which provides stronger coverage of biomedical terminology and syntax than domain-agnostic alternatives such as BERT-base [25] or SciBERT [26]. It has 768-dimensional sentence vectors and existing benchmarks report higher retrieval precision for biomedical queries compared with BioClinicalBERT [26] or BlueBERT [31].

In this case, the embedder will use the QAD to learn how to associate a clinician's free-text query (question) with a single, well-defined cohort description (answer), mirroring the real-world interaction and simplifying evaluation through exact-match.

Model fine-tuning was performed to adapt the embedder to the QA task. Specifically, we employed the MultipleNegativesRankingLoss (MNRL) [27] from the sentence-transformers library [26], a contrastive loss that, for every anchor–positive pair in a mini-batch, maximises their cosine similarity while treating all other instances in the batch as implicit negatives. This loss implements a cross-entropy objective known as InfoNCE [28], which compares the true (query, cohort) pair against the rest in the mini-batch by encouraging the model to assign a higher similarity to the correct pair. InfoNCE is a contrastive formulation originally designed to discriminate a true positive from a set of distractors by scaling the probability of correct matches. This in-batch negative sampling increases the number of effective negatives with batch size and has proved effective for retrieval tasks where explicit negatives are scarce [29,30].

Formally, for a query *q*, its *P* positive examples $p_i$ and the *N* in-batch negatives $n_j$, the loss is:

$$\text{Loss} = \sum_{i=1}^{P} \sum_{j=1}^{N} \max\left(0,\ f(q, p_i) - f(q, n_j) + \text{margin}\right)$$

where f() is the cosine-similarity function, *P* and *N* denote the counts of positive and negative pairs in the mini-batch, and *γ* is a margin hyperparameter that enforces the desired separation between positive and negative scores.

In our QAD, positive (Query, Cohort) pairs are explicitly labeled, while all other cohorts serve as implicit negatives. In comparison with the original release, we required fewer training epochs and warm-up steps due to the smaller size of our dataset (2 epochs, with warm-up covering 10% of total iterations).

**Step 4: embedding's evaluation**
Model performance was monitored via two complementary strategies in this order: (1) validation loss was measured every 5% of an epoch to detect underfitting or overtraining; (2) retrieval-based evaluation was performed using a ground truth query-to-cohort mapping (see Figure 1, Step 6). Using the trained embedder, cohorts were retrieved based on cosine similarity. For each query, the fine-tuned model returns the cohorts most similar to the query based on cosine similarity. To estimate how relevant were the answers of each query we use Retrieval Precision metric [22,23], computed as follows:

$$Precision = \frac{Number\ of\ Relevant\ Cohorts\ Retriever}{Total\ number\ of\ relevant\ cohorts}$$

Additionally, cohorts are ranked based on cosine similarity, which allows us to get the Mean Percentile Rank (MPR) for each query.

$$MPR = \frac{1}{N} \sum_{i=1}^{n} \frac{Rank\ of\ Cohort_i}{Total\ number\ of\ Cohorts}$$

where *N* is the number of correct cohorts for the query.

**Step 5: embeddings' deployment**
The resulting embedding model was deployed with Gradio's ChatInterface [32], which links a custom Python callback to a chat widget that accepts free-text biomedical queries. Each query is routed to the backend, the callback returns a markdown-formatted response, and Gradio streams the output to the browser. The Gradios's launch function creates a FastAPI server, serves the auto-generated HTML and JavaScript frontend, and lets the application run locally or be shared through a public URL without further web-development effort.

## Results
**Ontology-Augmented Normalization and Synonym Expansion Across Metadata Dimensions**
A total of 3823 omics cohorts were initially retrieved from GEO [1] across a range of NDs. After applying the disease-specific filtering protocol (see Methodology Step 2 Section), 2801 cohorts remained. The most represented condition was: AD (n = 1250); followed by (2) PD (n = 589) and HD (n = 365). Other conditions were less frequently represented, including LBD (n = 18) and MSA (n=17). In terms of heterogeneity of the terminology used to describe the entries at GEO, we identified 33 distinct population descriptors, 19 assay types, and 1770 non-standardized tissue annotations. For example, multiple distinct values were found to describe semantically equivalent values (e.g., "brain cortex" tissue vs. "cerebral cortex" tissue). All these terms need to be standardized in order to be used in downstream analysis.

To this end, first we homogenized raw values, i.e. every string was normalized and whenever a match existed (exact or similar), each value was replaced by the standard label of its reference ontology (see Section 2). Additionally, the Tissue field required normalization, i.e. consolidating those variants into a single canonical identifier so that semantically equivalent concepts are represented consistently across cohorts. From the original 1770 non-standardized entries retrieved from GEO, 560 values were successfully mapped to UBERON, MeSH and UMLS ontologies using direct and fuzzy matching strategies, 326 of them corresponding to unique standardized terms (see Table 1). Specifically, the majority of tissue mappings were obtained via fuzzy matching: 384 values (73.84%) compared to 136 (26.16%) obtained via exact matches. UBERON was the most informative source, contributing 227 standardized terms (43.7%). Then, for Po, As and Ph, an augmentation step was applied to enrich the GEO metadata. For Po, the initial set of 33 unique descriptors was augmented to a total of 105 synonym terms from the NCBI Taxonomy ontology. For As, 19 distinct values were augmented to 51 (only 11 of the initial values yielded synonyms). For Ph, we employed a combination of exact and fuzzy matching strategies, which led to 31 synonyms from EFO and 12 from UMLS. All 13 input values were successfully augmented to 292 values from the MeSH ontology, with 45 synonyms (15.4%) obtained from direct matches, and 247 (84.6%) inferred through fuzzy similarity scoring.

| Field | Original Count | No Match | Match | Synonyms | EFO/NCBI/UBERON (%) | MeSH (%) | UMLS (%) | Direct (%) | Fuzzy (%) | Final Count |
|---|---|---|---|---|---|---|---|---|---|---|
| Ti | 1700 | 1210 | 560 | 0 | 227 (43.65) | 87 (16.73) | 206 (39.62) | 136 (26.16) | 384 (73.84) | 326 |
| Po | 33 | 5 | 28 | 100 | 100 (100.0) | 0 (0.0) | 0 (0.0) | 100 (100.0) | 0 (0.0) | 105 |
| As | 19 | 8 | 11 | 43 | 31 (72.1) | 0 (0.0) | 12 (27.9) | 0 (0.0) | 43 (100.0) | 51 |
| Ph | 13 | 0 | 13 | 292 | 0 (0.0) | 292 (100.0) | 0 (0.0) | 45 (15.4) | 247 (84.6) | 292 |

**Table 1. Synonym-augmentation summary across Ti, Po, As and Ph dimensions**. The table reports: **(1)** the metadata field; **(2)** the number of original distinct terms retrieved from GEO (**Original Count**); **(3)** the number of those original distinct terms with no match in any ontology (**No Match**); **(4)** the number of matched terms (**Match**); **(5)** the total number of synonyms generated (**Synonyms**); **(6–8)** the distribution of synonym sources by ontology (**EFO/NCBI/UBERON** + **MeSH** + **UMLS**); and **(9–10)** the distribution of synonyms obtained via exact match (**Direct**) or fuzzy matching (**Fuzzy**).

**NLQ Dataset Construction Using Augmented Metadata**

To fine-tune the base embedder, we used the synonym-augmented metadata to construct a space of NLQs by combining values across the four standardized metadata dimensions. Since the expansion process resulted in 105 Po terms, 51 As terms, 292 Ph terms, and 326 Ti terms (774 unique values), the space of all combinations of single values for all four elements would lead to $5\times10^8$ NLQs. Nevertheless, we restricted this space to NLQs with verifiable answers (i.e., those with at least one matching cohort), leading to 368,082 NLQs, each one associated with 1.5 cohorts on average. Specifically, we partitioned the 774 augmented metadata values into training and test sets using metadata-stratified sampling, allocating 80% of the synonyms to training (n = 619) and 20% for evaluation (n = 155). This procedure led to two disjoint sets of NLQs: (1) NLQs composed exclusively of synonyms from the training subset (n = 139,336) and (2) NLQs composed exclusively of synonyms from the test subset (n = 1,886). To avoid overfitting caused by the large imbalance between the numbers of training-only and test-only NLQs, we randomly subsampled the training set to 7,544 NLQs, which is four times the size of the test set.

**Cohort Retrieval Fine-tuned Embedding Evaluation**

After training the base embedder with our NLQ-cohort dataset, we monitored both training and evaluation "MNRL" loss [33] to evaluate model convergence. We selected this loss function because it maximises the similarity margin between each true query-cohort pair and the many in-batch negatives, making it well suited for dense retrieval tasks (see Methodology Step 3 section, see Equation 1).

As shown in Figure 2A, the training loss decreased during the initial steps, from 1.10 at step 15 to 0.24 at step 60. Afterward, the training loss continued to decline more gradually and stabilized below 0.15 by step 255. The validation loss followed a similar trend, dropping from 0.33 at step 15 to approximately 0.12 from step 120 onward. The plateau below 0.15 shows that the encoder has already distilled the dominant patterns in the training NLQs. From that point on, each newly presented query mostly reinforces what the model already knows instead of uncovering new signals. These late-stage NLQs are not useless: they stabilise the weights and help prevent over-fitting to earlier batches. Meaningful gains would require examples with NLQs with lexical variants, different metadata combinations or harder negative cohorts.

To assess the retrieval capabilities of the embedding model, we evaluated performance on the test set NLQs using two embedder metrics: Retrieval Precision and MPR (see Methodology Step 4 Section). As shown in Figure 2B, For each group of NLQs, defined by the number of metadata terms included in the query (*n_terms*, i.e., 1, 2, 3, or 4 of the dimensions Po, As, Ph, or Ti), the majority of instances fell near 1.0 precision. Focusing first on two-term NLQs, we identified two distinct patterns. A small subset recorded precision values near 0.0 yet still retrieved the correct cohort within the top four results (mean MPR = 0.267), indicating that although competing partial matches scored slightly higher, the target cohort remained highly ranked. The remaining two-term queries showed precisions between 0.5 and 1.0 (mean MPR = 0.834). For three-term NLQs, nearly all cases exceeded 0.5 precision; only a few targeted a single cohort with precision below 0.5, and their mean MPR was 0.285, so the correct match was still placed in the top four despite one atypical metadata term. Finally, all four-term NLQs clustered at 1.0 precision. We also observed that all test NLQs, regardless of the number of terms, achieved MPR values near 1.0, following a similar pattern to retrieval precision (see Figure 2C). However, unlike retrieval precision, MPR showed no accumulation of NLQs near 0.0, suggesting that even when retrieval precision is zero, the correct result is still ranked relatively high by the embedding model.

Finally, to evaluate the impact of fine-tuning, we compared the retrieval performance of the trained embedder to that of the original, non-fine-tuned "NeuML/pubmedbert-base-embeddings" model [34]. We detected a substantially lower performance of the base embedder compared to the fine-tuned using both metrics, retrieval precision and MPR. Specifically, a large number of queries had precision close to 0.0 (see Figure 3A), which indicates that, for the base model, relevant cohorts frequently appeared in the bottom percentiles of the ranked list (see Figure 3B).

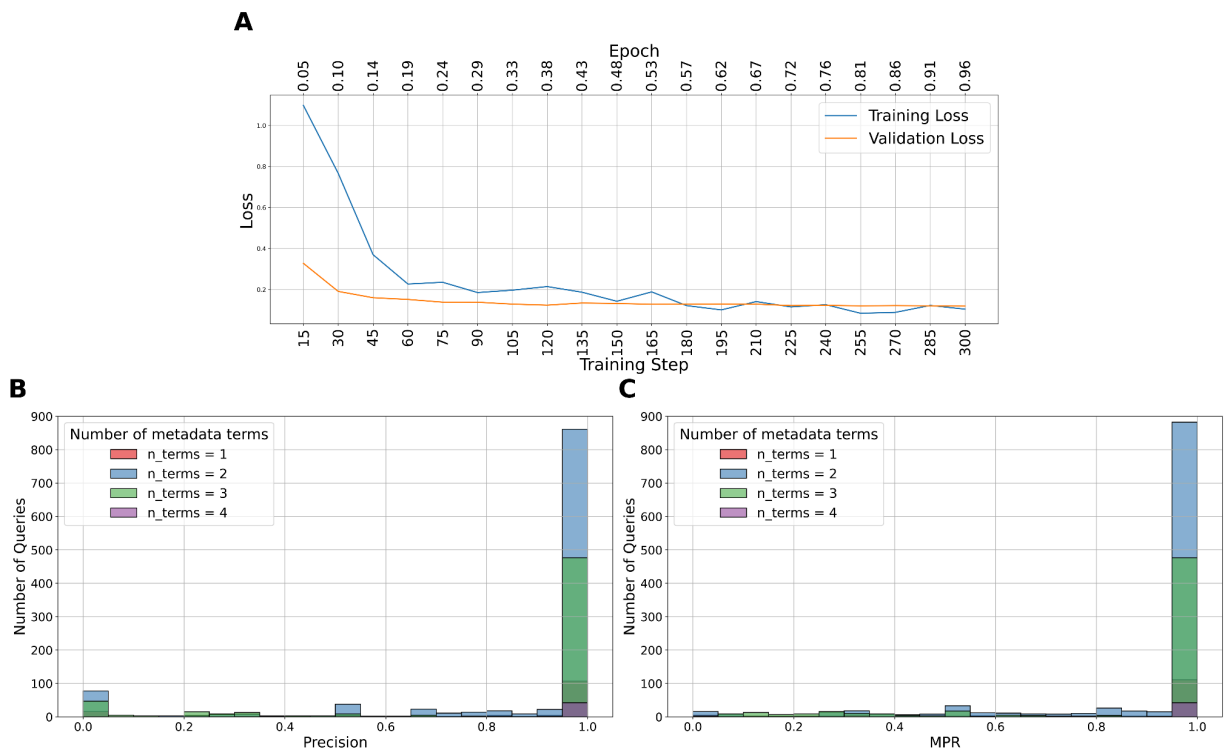

**Figure 2. (A)** Training and evaluation loss curves during fine-tuning. **(B)** Precision and **(C)** MPR distributions for test NLQs grouped by number of metadata terms using the fine-tuned embedder.

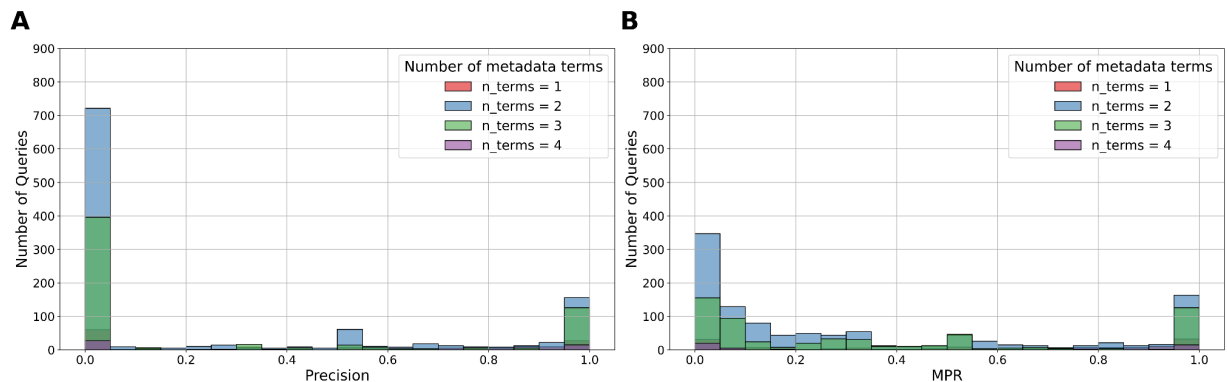

**Figure 3.** (A) Precision and (B) MPR distributions for test NLQs grouped by number of metadata terms using the base embedder.

## Discussion

We have defined an easy and adaptable methodology to create semantically rich repositories of omics cohorts based on LLMs as embedders. To this end, we use established ontologies to normalize structureless descriptions at the repositories and then we augment these descriptions through the use of synonyms, which will be used to fine-tune the embedder. That is, this workflow combines cohort metadata curation, large-scale synonym expansion across biomedical ontologies and fine-tuning of a PubMedBERT base encoder. Finally, we illustrated the applicability of this workflow for searching and discovering new NDs cohorts of the GEO repository.

Our experiments demonstrated that enriching the four core metadata dimensions (i.e., Po, As, Ph and Ti) with ontology-derived synonyms triples the lexical coverage of Po descriptors and expands Ph terms twenty times (see Table 1). When these curated descriptors are included in a shared embedding semantic space, the model attains a mean Retrieval Precision of 0.866 and a MPR of 0.896 on 1,886 natural-language queries, far surpassing the baseline PubMedBERT encoder (Retrieval Precision = 0.277; MPR = 0.355). Notably, queries that combine all four metadata dimensions achieve perfect Retrieval Precision. Moreover, error analysis reveals that most residual failures concentrate in two-term queries containing potentially ambiguous combinations (e.g., "Macaca Mulatta" with "Macaca Fascicularis" or "Bos Indicus" with "Bos Taurus"), suggesting that further improvements will stem from ontological disambiguation and expanded semantic coverage rather than from modifying the embedding architecture.

In the interactive Gradio-based platform, researchers can formulate free-text questions such as "Show me Parkinson's disease cohorts profiled with RNA-Seq in substantia nigra tissue" and receive highly relevant studies without laborious manual filtering. The interface returns a concise ranked list in plain text where each entry shows the cohort title, GEO accession and all the metadata associated with the cohort together with a quick link to the original GEO record. The application runs entirely in the browser, requires no local installation, which makes the retrieval workflow readily accessible to researchers with minimal technical effort.

This work also lays the foundation for a much broader initiative. We are currently generalizing the approach beyond the four metadata dimensions and cohort-level granularity. Specifically, we are working to index every individual ND samples (>150,000) contained in the same GEO studies together with all of their metadata dimensions. In the future, we intend to integrate GEO omics profiles so that similarity can be computed jointly over structured descriptors and latent molecular signatures. Our long-term objective is to support compound queries such as "Return two mouse samples exhibiting Parkinsonian phenotypes with the most similar multi-omics fingerprints", where similarity will be evaluated within an embedded vector space that fuses clinical terms and omic feature representations.